\documentclass[10pt,twocolumn,letterpaper]{article}

\usepackage[pagenumbers]{cvpr} 

\usepackage{graphicx}
\usepackage{amsmath}
\usepackage{amssymb}
\usepackage{booktabs}
\usepackage{tabularx}
\usepackage{multirow}
\usepackage[dvipsnames]{xcolor}
\usepackage[accsupp]{axessibility}
\usepackage[pagebackref,breaklinks,colorlinks]{hyperref}
\usepackage[normalem]{ulem}
\usepackage[accsupp]{axessibility}

\usepackage[capitalize]{cleveref}
\crefname{section}{Sec.}{Secs.}
\Crefname{section}{Section}{Sections}
\Crefname{table}{Table}{Tables}
\crefname{table}{Tab.}{Tabs.}

\newcommand{\remark}[1]{#1}
\newcommand{\ignore}[1]{}
\newcommand{\lj}[1]{}

\def\cvprPaperID{1890} 
\def\confName{CVPR}
\def\confYear{2023}

\begin{document}

\title{3D Video Loops from Asynchronous Input}

\author{
Li Ma$^{1}$ \qquad Xiaoyu Li$^{2}$ \qquad Jing Liao$^{3}$ \qquad Pedro V. Sander$^{1}$ 
\vspace{5pt}\\
$^{1}$The Hong Kong University of Science and Technology \\ $^{2}$Tencent AI Lab \qquad $^{3}$City University of Hong Kong \\
}

\twocolumn[{%
\renewcommand\twocolumn[1][]{#1}%
\maketitle

\begin{center}
    \centering
\captionsetup{type=figure}
\vspace{-0.5cm}
  \includegraphics[width=\linewidth]{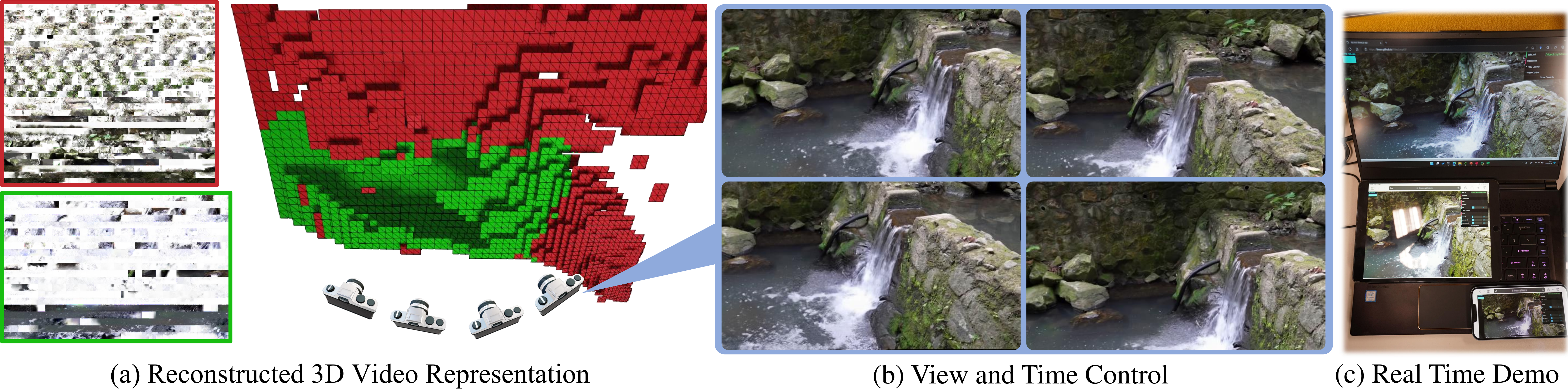}
  \caption{Given a set of asynchronous multi-view videos, we propose a pipeline to construct a novel 3D looping video representation (a), which consists of a {\color{Red}static texture atlas}, a {\color{Green}dynamic texture atlas}, and multiple tiles as the geometry proxy. The 3D video loops allow both view and time control (b), and can be rendered in real time even on mobile devices (c). We strongly recommend readers refer to the \textit{supplementary material} for video results.\lj{it is better to have larger view differences in (b)}}
  \label{fig:teaser}
\end{center}%
}]


\begin{abstract}
\vspace{-0.7cm}
Looping videos are short video clips that can be looped endlessly without visible seams or artifacts. They provide a very attractive way to capture the dynamism of natural scenes. Existing methods have been mostly limited to 2D representations. In this paper, we take a step forward and propose a practical solution that enables an immersive experience on dynamic 3D looping scenes. The key challenge is to consider the per-view looping conditions from asynchronous input while maintaining view consistency for the 3D representation. We propose a novel sparse 3D video representation, namely Multi-Tile Video (MTV), which not only provides a view-consistent prior, but also greatly reduces memory usage, making the optimization of a 4D volume tractable. Then, we introduce a two-stage pipeline to construct the 3D looping MTV from completely asynchronous multi-view videos with no time overlap. A novel looping loss based on video temporal retargeting algorithms is adopted during the optimization to loop the 3D scene. Experiments of our framework have shown promise in successfully generating and rendering photorealistic 3D looping videos in real time even on mobile devices. The code, dataset, and live demos are available in \url{https://limacv.github.io/VideoLoop3D_web/}.
\end{abstract}


\section{Introduction}
Endless looping videos are fascinating ways to record special moments. These video loops are compact in terms of storage and provide a much richer experience for scenes that exhibit looping behavior. One successful commercial use of this technique is the live photo \cite{misc_livephoto} feature in the Apple iPhone, which tries to find an optimal looping period and fade in/out short video clips to create looping videos. 
There have been several works on automatically constructing 2D looping videos from non-looping short video clips. Liao et al. \cite{loop_vid_progressive} first propose to create 2D video loops from videos captured with static cameras. They solve for the optimal starting frame and looping period for each pixel in the input video to composite the final video. Later on, several methods are proposed to improve the computation speed \cite{loop_vid_fast}, or extend to panoramas \cite{loop_vid_panoramic,loop_vid_dynamosaics}, and gigapixel videos \cite{loop_vid_gigapixel}. 
\remark{However, few attempts have been made to extend video loops to a 3D representation.
One existing work that shares a similar setting as ours is VBR \cite{VBR_loop}, which generates plausible video loops in novel views. However, it comes with some limitations: It builds on top of ULR \cite{ULR}, which can produce ghosting artifacts due to inaccurate mesh reconstruction, as shown in \cite{MPI_LLFF}. Besides, VBR generates looping videos and reduces the inconsistency from asynchronous input by adaptively blending in different frequency domains, which tends to blur away details. 
}
\ignore{However, to the best of our knowledge, no attempt has been made to extend video loops to a 3D representation. Such an extension is a non-trivial problem that remains to be explored.}

To allow free-view observation of the looping videos, a proper 3D representation needs to be employed. Recently, tremendous progress has been made in novel view synthesis based on 3D scene representations such as triangle meshes \cite{mesh_neuraltex,mesh_freevs,mesh_stablevs}, Multi-plane Image (MPI) \cite{MPI_stereomag,MPI_deepview}, and Neural Radiance Field (NeRF) \cite{NeRF,NGP,TensorRF}, which could be reconstructed given only sparse observations of real scenes and render photo-realistic images in novel views. Much effort has been made to adapt these methods to dynamic scenes, which allows for both viewing space and time controls~\cite{dynNVS_microsoft,volume_NV,volume_MVP,NeRFrgbddyn_NRD,NeRFrgbddyn_tofnerf,svdyn_yoon,NeRFsvdyn_nerfies,NeRFsvdyn_HyperNeRF}. Therefore, a straightforward solution to generate a 3D looping video is to employ the 2D looping algorithms for each view and lift the results to 3D using these methods. However, we find it hard to get satisfactory results since the 2D looping algorithms do not consider view consistency, which is even more challenging for the asynchronous multi-view videos that we use as input.

In this work, we develop a practical solution for these problems by using the captured video input of the dynamic 3D scene with only one commodity camera. We automatically construct a 3D looping video representation from completely asynchronous multi-view input videos with no time overlap. To get promising 3D video loop results, two main issues need to be addressed. First, we need to solve for a view-consistent looping pattern from inconsistent multi-view videos, from which we need to identify spatio-temporal 3D patches that are as consistent as possible. Second, the 3D video potentially requires a memory-intensive 4D volume for storage. Therefore, we need to develop a 3D video representation that is both efficient in rendering and compact in memory usage to make the optimization of the 4D volume tractable. 

To this end, we develop an analysis-by-synthesis approach that trains for a view-consistent 3D video representation by optimizing multi-view looping targets. We propose an efficient 3D video representation based on Multi-plane Images (MPIs), namely Multi-tile Videos (MTVs), by exploiting the spatial and temporal sparsity of the 3D scene. As shown in Fig.~\ref{fig:repr}, instead of densely storing large planes, MTVs store static or dynamic texture tiles that are sparsely scattered in the view frustum. This greatly reduces the memory requirement for rendering compared with other 3D video representations, making the optimization of the 3D looping video feasible in a single GPU. The sparsity of MTVs also serves as a view-consistent prior when optimizing the 3D looping video.
To optimize the representation for looping, we formulate the looping generation for each view as a temporal video retargeting problem and develop a novel looping loss based on this formulation. We propose a two-stage pipeline to generate a looping MTV, and the experiments show that our method can produce photorealistic 3D video loops that maintain similar dynamism from the input, and enable real-time rendering even in mobile devices. Our contributions can be summarized as follows:
\begin{itemize}
    \vspace{-0.1cm}
    \item We propose Multi-tile Videos (MTVs), a novel dynamic 3D scene representation that is efficient in rendering and compact in memory usage.
    \vspace{-0.1cm}
    \item We propose a novel looping loss by formulating the 3D video looping construction as a temporal retargeting problem. 
    \vspace{-0.1cm}
    \item We propose a two-stage pipeline that constructs MTVs from completely asynchronous multi-view videos.
\end{itemize}

\section{Related Work}
Our work lies at the confluence of two research topics: looping video construction and novel view synthesis. We will review each of them in this section.

\vspace{0.2cm}
\noindent\textbf{Video Loops.}
Several works have been proposed to synthesize looping videos from short video clips. Sch{\"o}dl et al. \cite{loop_vid_videotex} create video loops by finding similar video frames and jumping between them. Audio \cite{audio} can also be leveraged for further refinement. Liao et al. \cite{loop_vid_progressive} formulate the looping as a combinatorial optimization problem that tries to find the optimal start frame and looping period for each pixel. It seeks to maximize spatio-temporal consistency in the output looping videos. This formulation is further developed and accelerated by Liao et al. \cite{loop_vid_fast}, and extended to gigapixel looping videos \cite{loop_vid_gigapixel} by stitching multiple looping videos. Panorama video loops can also be created by taking a video with a panning camera \cite{loop_vid_panoramic,loop_vid_dynamosaics}.  
\remark{VBR \cite{VBR_loop} generates loops by fading in/out temporal Laplacian pyramids, and extends video loops to 3D using ULR \cite{ULR}. }
Another line of work tries to create video loops from still images and strokes provided by users as rough guidelines of the looping motion. Endless Loops \cite{loop_img_endlessLoops} tries to find self-similarities from the image and solve for the optical flow field, which is then used to warp and composite the frames of the looping video. This process can also be replaced by data-driven approaches \cite{loop_img_controllable,loop_img_eulerian}\remark{, or physics-based simulation \cite{loop_img_simulatefluid}.}
Despite the progress in creating various forms of looping videos, extending looping videos to 3D is still an unexplored direction.


\vspace{0.2cm}
\noindent\textbf{Novel View Synthesis of Dynamic Scenes.}
Novel View Synthesis (NVS) aims at interpolating views given only a set of sparse input views. For dynamic scenes, NVS requires the construction of a 4D representation that allows for both space and time control. Some methods use synchronized multi-view videos as input, which are often only available in a studio setting \cite{dynNVS_microsoft,volume_NV,volume_MVP}, or using specially designed camera arrays \cite{MPI_immersive,NeRFmvdyn_DynNeRF,MPI_deepview,MPI_dynmask}. 
To ease hardware requirements, Open4D \cite{dynNVS_open4d} uses unconstrained multi-view input, but still requires multiple observations at the same timestamp. With the development of neural rendering, it is possible to use only monocular input. However, this is a highly ill-posed problem since the camera and scene elements are moving simultaneously. Some methods use extra sensors such as a depth sensor \cite{NeRFrgbddyn_tofnerf,NeRFrgbddyn_NRD}, while some use a data-driven prior to help construct the scene geometry \cite{NeRFsvdyn_3Dmoments,NeRFsvdyn_dynamicNVS}. Others use a hand-crafted motion prior to regularize the scene motion \cite{NeRFsvdyn_nerfies,NeRFsvdyn_nonrigid,NeRFsvdyn_HyperNeRF,NeRFsvdyn_NSFF}, which usually can only handle simple motions. In our setting, we take asynchronous multi-view videos with no time overlap, which is a setting that has not been addressed before. 

\vspace{0.1cm}
\noindent\textbf{3D Scene Representations. }
A critical issue in NVS is the underlying scene representation. A triangle mesh is the most commonly used scene representation in commercial 3D software. Some methods use meshes as their representation \cite{VBR_loop,mesh_stablevs,mesh_freevs}. However, reconstructing an accurate, temporally consistent mesh is still an open problem, being particularly challenging for complex in-the-wild scenes \cite{volume_MVP}. A volumetric representation is another option to express the 3D world by storing scene parameters in a dense 3D grid \cite{volume_svox2,volume_NV,volume_NeuVV,volume_fourieroctree}. One benefit is that it trivially supports differentiable rendering, which greatly improves the reconstruction quality. The Multi-plane Image (MPI) \cite{MPI_deepstereo,MPI_LLFF,MPI_pushbound,MPI_deepview,MPI_stereomag,MPI_svmpi} is an adapted volumetric representation that represents a scene using multiple RGBA planes in the camera frustum. Volume representations can model complex geometry, but at the cost of higher memory usage. Another rapidly developing representation is Neural Radiance Field (NeRF) \cite{NeRF}, which models scenes as continuous functions and parameterizes the function as an implicit neural network. It achieves photorealistic rendering results at the expense of long training and rendering times, especially for dynamic scenes. 


\section{Method}
\subsection{Overview}
\begin{figure*}
    \centering
    \includegraphics[width=0.98\linewidth]{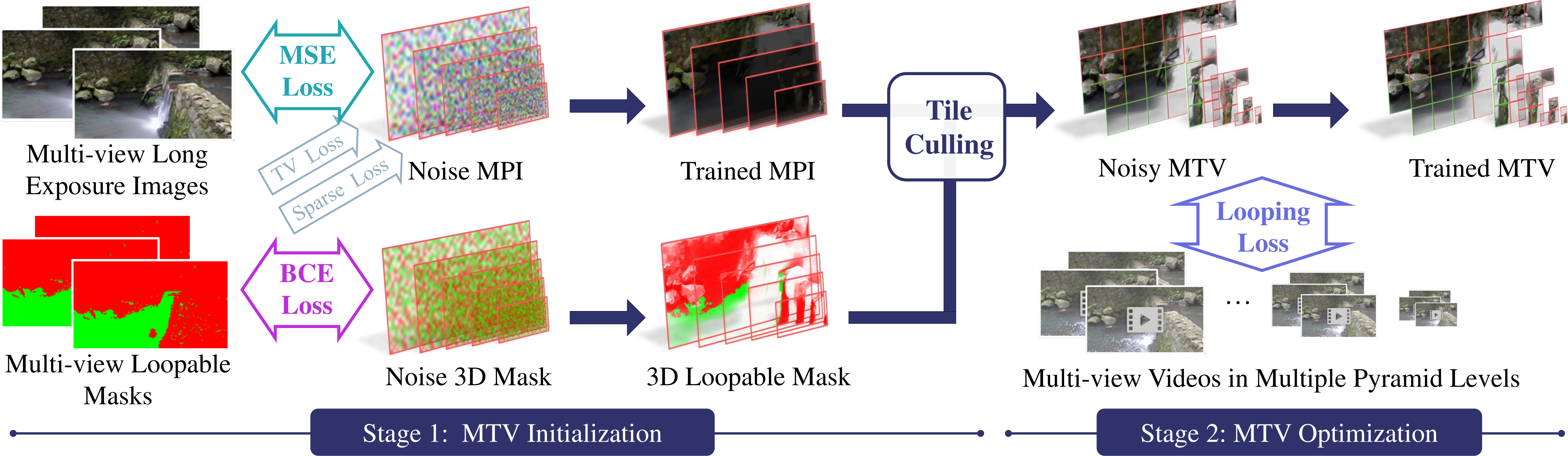}
    \vspace{-0.2cm}
    \caption{The two-stage pipeline to generate the MTV representation from multi-view videos.}
    \label{fig:pipeline}
    \vspace{-0.4cm}
\end{figure*}

Our goal is to reconstruct a view-consistent 3D video representation that can be looped infinitely using completely asynchronous multi-view 2D videos. 
We start by introducing a novel 3D video representation, namely Multi-tile Videos (MTVs), which improves efficiency by exploiting sparsity. Then we propose a two-stage pipeline as shown in Fig.~\ref{fig:pipeline} to construct a 3D looping MTV. In the first stage, we initialize the MTV by optimizing a static Multi-plane Image (MPI) and a 3D loopable mask using long exposure images and 2D loopable masks derived from the input videos. We then construct an MTV through a tile culling process. In the second stage, we train the MTV using an analysis-by-synthesis approach in a coarse-to-fine manner. The key enabler for this process is a novel looping loss based on video retargeting algorithms, which encourages a video to simultaneously loop and preserve similarity to the input. The remainder of this section describes the details of this proposed approach.

\begin{figure}
    \centering
    \includegraphics[width=\linewidth]{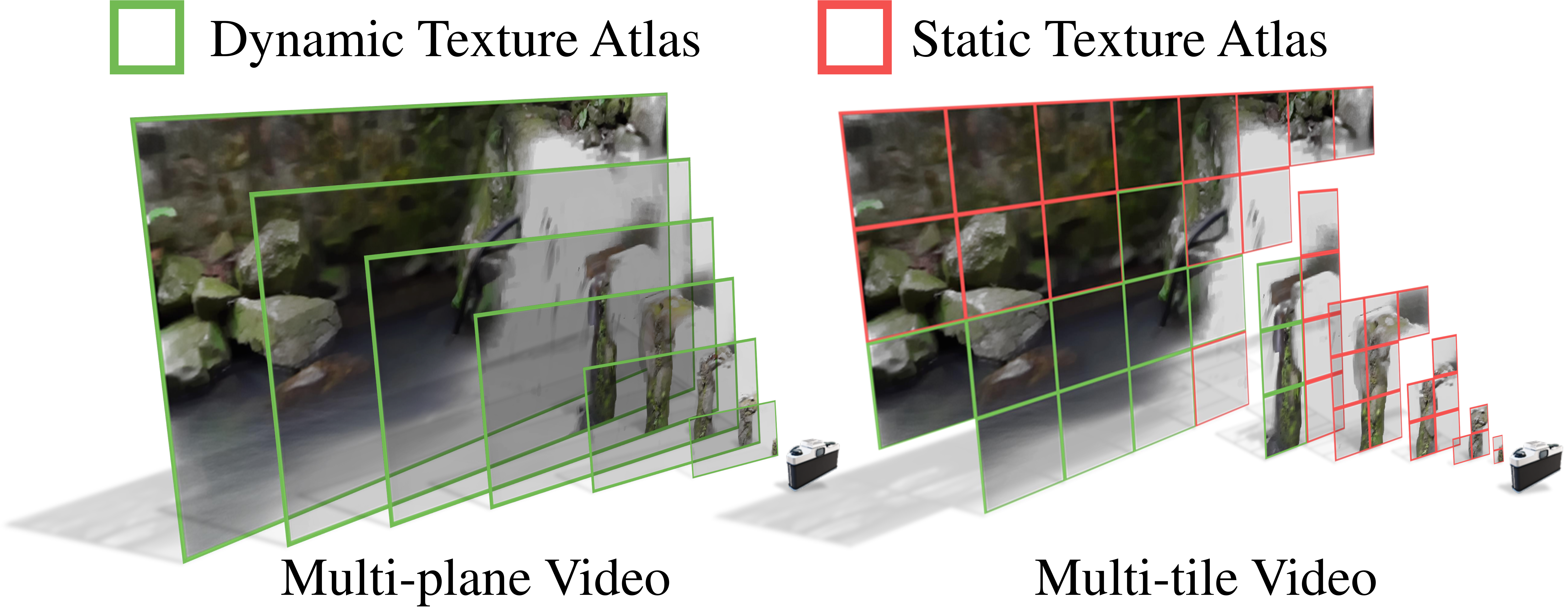}
    \vspace{-0.4cm}
    \caption{Comparison between the Multi-plane Video representation and the Multi-tile Video representation. }
    \label{fig:repr}
    \vspace{-0.3cm}
\end{figure}


\subsection{Data Preparation}
The input to our system are multiple asynchronous videos of the same scene from different views. Each video $\mathbf{V} \in \mathbb{R}^{F \times H \times W \times 3}$ is a short clip with $F$ frames and a resolution of $H \times W$. Video lengths may differ for each view. Each video is expected to have a fixed camera pose, which can be achieved using tripods or existing video stabilization tools during post-process. Since we allow videos to be asynchronous, we could capture each view sequentially using a single commodity camera. 

Given the precondition that the captured scene contains mostly repetitive content, we assume the long exposure images for each view to be view-consistent. Therefore, we compute an average image 
for each video $\mathbf{V}$, and then register a pinhole camera model for each video using COLMAP \cite{misc_colmap1,misc_colmap2}. We also compute a binary loopable mask 
for each input video similar to Liao et al. \cite{loop_vid_fast}, where $1$ indicates pixel with the potential to form a loop and $0$ otherwise.

\subsection{Multi-tile Video (MTV) Representation}
Before introducing our proposed MTV representation, we first briefly review the MPI representation \cite{MPI_stereomag}. An MPI represents the scene using $D$ fronto-parallel RGBA planes in the frustum of a reference camera, with each plane arranged at fixed depths \cite{MPI_svmpi}.
To render an MPI from novel views, we first need to warp each plane based on the depth of the plane and the viewing camera, and then iteratively blend each warped plane from back to front. A straightforward dynamic extension of MPI, namely Multi-plane Video (MPV), is to store a sequence of RGBA maps for each plane. For a video with $T$ frames, this results in a 4D volume in $\mathbb{R} ^ {D \times T \times H \times W \times 4}$, which is very memory consuming. Inspired by recent work on sparse volume representation \cite{sparse_baking,sparse_NSVF,sparse_plenoctree}, we propose Multi-tile Videos, which reduce the memory requirements by exploiting the spatio-temporal sparsity of the scene. Specifically, we subdivide each plane into a regular grid of tiny tiles. Each tile $\mathbf{T} \in \mathbb{R} ^ {F \times H_{s} \times W_{s} \times 4}$ stores a small RGBA patch sequence with spatial resolution $H_{s} \times W_{s}$. For each tile, we assign a label $l$ by identifying whether it contains looping content $l_{loop}$, a static scene $l_{static}$, or is simply empty $l_{empty}$. We could then store a single RGBA patch for $l_{static}$, and discard tiles that are empty. Fig.~\ref{fig:teaser} visualizes a reconstructed MTV representation, where the RGBA patches are packed into {\color{Red}static} and {\color{Green}dynamic} texture atlas. Fig.~\ref{fig:repr} shows the difference between MPVs and MTVs.

\subsection{Stage 1: MTV Initialization}
We find that optimizing a dense MTV directly from scratch results in the approach being easily trapped in local minima, which yields view-inconsistent results. To address this, we use a two-stage pipeline shown in Fig.~\ref{fig:pipeline}. In the first stage, we start by constructing a ``long exposure'' MPI. Then we initialize the sparse MTV by tile culling process that removes unnecessary tiles. By reducing the number of parameters, the initialized MTV provides a view-consistent prior and leads to a high-quality 3D video representation.

\paragraph{Training a looping-aware MPI.}
We start by training a dense MPI $\mathbf{M} \in \mathbb{R}^{D \times H \times W \times 4}$, as well as a 3D loopable mask $\mathbf{L} \in \mathbb{R}^{D \times H \times W}$, using the average image and the 2D loopable mask, respectively. We randomly initialize $\mathbf{M}$ and $\mathbf{L}$, and in each iteration, we randomly sample a patch in a random view, and render an RGB patch $\mathbf{\hat{p}}_{c} \in \mathbb{R}^{h \times w \times 3}$ and a loopable mask patch $\mathbf{\hat{p}}_{l} \in \mathbb{R}^{h \times w}$ using the standard MPI rendering method. Note that the $\alpha$ channel is shared between $\mathbf{M}$ and $\mathbf{L}$ during rendering. We supervise the MPI $\mathbf{M}$ by minimizing the Mean Square Error (MSE) between the rendering results and the corresponding patch $\mathbf{p}_{c}$ from the average image. We supervise the loopable mask $\mathbf{L}$ by minimizing the Binary Cross Entropy (BCE) between the rendered 2D mask $\mathbf{\hat{p}}_{l}$ and the corresponding patch $\mathbf{p}_{l}$ from the 2D loopable mask:
\begin{align}
    \mathcal{L}_{mse} &= \frac{1}{h w} \|\mathbf{p}_{c} - \mathbf{\hat{p}}_{c}\|_2^2, \\
    \mathcal{L}_{bcd} &= \frac{1}{h w} \| -(\mathbf{p}_{l} log(\mathbf{\hat{p}}_{l}) + (1 - \mathbf{p}_{l}) log(   1 - \mathbf{\hat{p}}_{l})) \|_1,
\end{align}
where $\|\mathbf{p}\|_1$ and $\|\mathbf{p}\|_2$ are the L1 and L2 norm of a flattened patch $\mathbf{p}$. The $log$ is computed for every element of a patch. Since the rendering of the MPI is differentiable, we optimize $\mathbf{M}$ and $\mathbf{L}$ using the Adam optimizer \cite{misc_adam}.
Optimizing all the parameters freely causes noisy artifacts, therefore, we apply total variation (TV) regularization \cite{misc_TV} to $\mathbf{M}$:
\begin{equation}
    \mathcal{L}_{tv} = \frac{1}{H W} (\|\Delta_x \mathbf{M}\|_1 + \|\Delta_y \mathbf{M}\|_1),
\end{equation}
where $\|\Delta_x \mathbf{M}\|_1$ is shorthand for the L1 norm of the gradient of each pixel in the MPI $\mathbf{M}$ along $x$ direction, and analogously for $\|\Delta_y \mathbf{M}\|_1$. We also adopt a sparsity loss to further encourage sparsity to the $\alpha$ channel of the MPI $\mathbf{M}_\alpha$ as in Broxton et al. \cite{MPI_immersive}. Specifically, we collect $D$ alpha values in each pixel location of $\mathbf{M}_\alpha$ into a vector $\mathbf{\beta}$, where $D$ is the number of planes. Then the sparsity loss is computed as:
\begin{equation}
    \mathcal{L}_{spa} = \frac{1}{H W} \sum_{pixel} \frac{\|\mathbf{\beta}\|_1}{\|\mathbf{\beta}\|_2}.
\end{equation}
The final loss in the first stage is a weighted sum of the four losses:
\begin{equation}
    \mathcal{L} = \mathcal{L}_{mse} + \mathcal{L}_{bcd} + \lambda_{tv} \mathcal{L}_{tv} + \lambda_{spa} \mathcal{L}_{spa}.
\end{equation}

\paragraph{Tile Culling. }
After training, we reconstruct a static MPI $\mathbf{M}$ as well as a 3D loopable mask $\mathbf{L}$. We subdivide each plane into a regular grid of tiles. In the experiments, we subdivide the plane so that each tile has a resolution of $H_s = W_s = 16$. We denote $\{T_c\}$, $\{T_\alpha\}$, $\{T_l\}$ to be the set of RGB color, alpha value, and loopable mask of a tile, respectively. We then assign label $l \in \{l_{empty}, l_{static}, l_{loop}\}$ based on the $\{T_\alpha\}$ and $\{T_l\}$ for each tile: 
\begin{equation}
    l = 
    \begin{cases}
    l_{empty} & \text{if } max\{T_\alpha\} \leq \tau_\alpha \text{, }  \\
    l_{static} & \text{if } max\{T_\alpha\} > \tau_\alpha \text{and } max\{T_l\} < \tau_l \text{, }\\
    l_{loop} & \text{otherwise.}
    \end{cases}
\end{equation}
We set the threshold of culling to be $\tau_\alpha = 0.05$ and $\tau_l = 0.5$. We cull the tiles with $l = l_{empty}$. For tiles with $l = l_{loop}$, we lift the static 2D RGBA patch into a patch sequence by copying the patch $T$ times, where $T$ is the number of frames that we would like the MTV to have. We add a small random noise to the lifted patch video to prevent the straightforward static solution. For tiles with $l = l_{static}$, we simply keep it unchanged. This culling process greatly reduces the memory requirement for optimizing the 4D volume.

\subsection{Stage 2: MTV Optimization}
After initializing the MTV representation, we then seek to optimize the final looping MTV. 

\begin{figure}
    \centering
    \includegraphics[width=\linewidth]{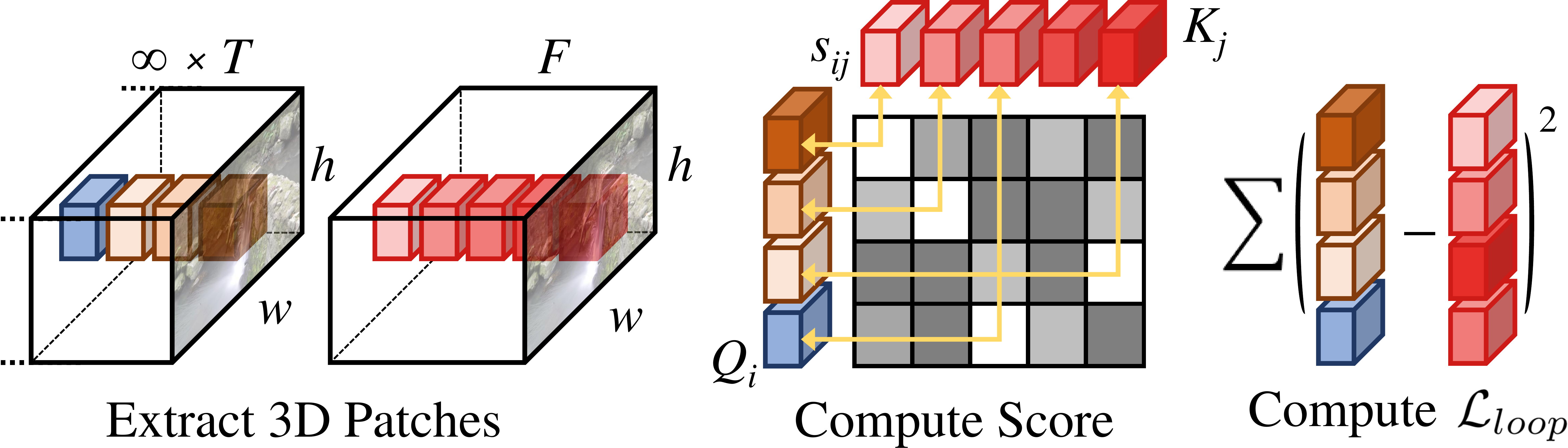}
    \caption{Visualization of looping loss. We first pad frames and extract 3D patches along the time axis for each pixel location, then we compute a normalized similarity score for each patch pair. Finally, the looping loss is computed by averaging errors between patches with minimum scores.}
    \label{fig:loop}
    \vspace{-0.4cm}
\end{figure}

\paragraph{Looping Loss.}
The main supervision of the optimization process is a novel looping loss, which is inspired by the recent progress in image \cite{retarget_GPNN} and video \cite{retarget_vgpnn} retargeting algorithm. Specifically, in each iteration, we randomly sample a view and a rectangle window of size $h \times w$, and render the video $\mathbf{\hat{V}}_o \in \mathbb{R}^{T \times h \times w \times 3}$ from MTV. We denote the corresponding input video as $\mathbf{V}_p \in \mathbb{R}^{F \times h \times w \times 3}$. Our goal is to optimize the MTV such that $\mathbf{\hat{V}}_o$ forms a looping video $\mathbf{V}_\infty$: 
\begin{equation}
    \mathbf{V}_\infty(t) = \mathbf{\hat{V}}_o(t \text{ mod } T), t \in [1, +\infty),
\end{equation}
where $\mathbf{V}(t)$ means $t$-th frame of the video and mod is the modulus operation. We define the looping loss to encourage the $\mathbf{V}_\infty$ to be a temporal retargeting result of $\mathbf{V}_p$. A visualization of the process is shown in Fig.~\ref{fig:loop}. 

\begin{figure}
    \centering
    \includegraphics[width=0.98\linewidth]{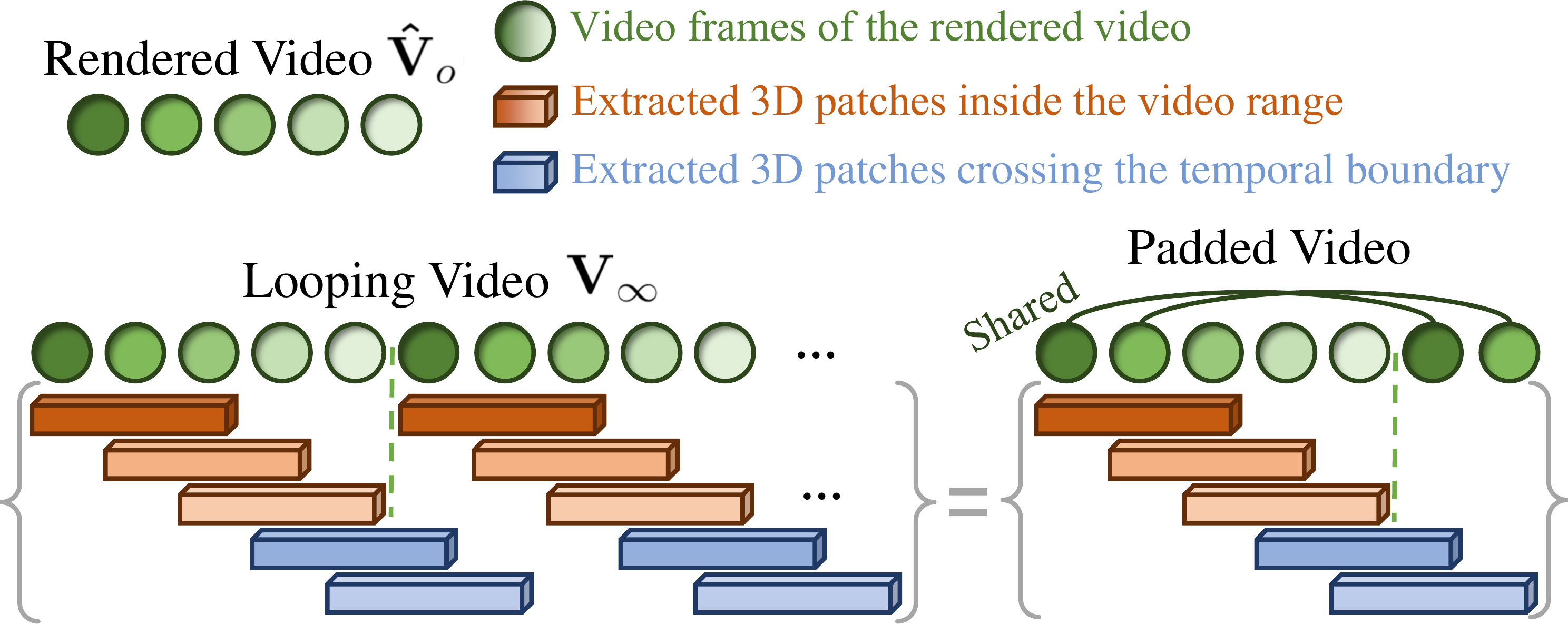}
    \caption{For patches of size $3$ and stride $1$, the patch set extracted from the video that endlessly repeats $5$ frames is the same as the patch set extracted from the padded video that circularly pads $2$ frames. }
    \vspace{-0.4cm}
    \label{fig:pad}
\end{figure}

We start by extracting 3D patch sets $\{\mathbf{Q}_i; i = 1,...,n\}$ and $\{\mathbf{K}_j; j = 1,...,m\}$ from $\mathbf{V}_\infty$ and $\mathbf{V}_p$, respectively, along temporal axis. $\{\mathbf{Q}_i\}$ and $\{\mathbf{K}_j\}$ are all centered at the same pixel location and we repeat the same process for every pixel. Note that although there are infinitely many patches from the looping video, the extracted patch set of the looping video is equivalent to a finite set of patches, which are extracted from the rendered video by circularly padding the first $p = s - d$ frames of the rendered video $\mathbf{\hat{V}}_o$ at the end of itself, where $s$ and $d$ are the size and stride of the patches in the time axis. Fig.~\ref{fig:pad} demonstrates a toy example with $5$ frames. By optimizing both the patches inside the video range and patches crossing the temporal boundary, we optimize a video that is both spatio-temporally consistent with the target and seamlessly looping. We then try to minimize the bidirectional similarity (BDS) \cite{retarget_bds} between the two sets of patches. Intuitively, this means every patch in $\{\mathbf{Q}_i\}$ appears in $\{\mathbf{K}_j\}$ (for coherence) and every patch in $\{\mathbf{K}_j\}$ appears in $\{\mathbf{Q}_i\}$ (for completeness).

To minimize the BDS between the two patch sets, we use the Patch Nearest Neighbor (PNN) algorithm \cite{retarget_GPNN} that first computes a 2D table of normalized similarity scores (NSSs) $s_{i j}$ for every possible pair of $\mathbf{Q}_i$ and $\mathbf{K}_j$. Then for each patch $\mathbf{Q}_i$, we select a target patch $\mathbf{K}_{f(i)} \in \{\mathbf{K}_j\}$ that has minimal NSS, where $f(i)$ is a selection function:
\begin{align}
    f(i) &= \arg\min_{k} s_{i, k}, \text{where} \\
    s_{i j} &= \frac{1}{\rho + \min_k \|\mathbf{Q}_k - \mathbf{K}_j\|_2^2} \|\mathbf{Q}_i - \mathbf{K}_j\|_2^2.
    \label{eq:sij}
\end{align}
Here $\rho$ is a hyperparameter that controls the degree of completeness. Intuitively, when $\rho \rightarrow \inf$, Eq.~\ref{eq:sij} degenerates to $s_{ij} \sim D(\mathbf{Q}_i, \mathbf{K}_j)$, so we simply select $\mathbf{K}_j$ that is most similar to $\mathbf{Q}_i$. And if $\rho = 0$, the denominator $\min_k D(\mathbf{Q}_k, \mathbf{K}_j)$ penalizes the score if there are already some $\mathbf{Q}_i$ that is closest to $\mathbf{K}_j$. Thus, the selection will prefer patches that have not yet been selected.

Using the PNN algorithm, we get the set of patches $\{\mathbf{K}_{f(i)}\}$ that is coherent to the target patch set $\{\mathbf{K}_j\}$, and the completeness is controlled by $\rho$. The looping loss is then defined as the MSE loss between $\mathbf{Q}_i$ and $\mathbf{K}_{f(i)}$:
\begin{equation}
    \mathcal{L}_{loop} = \frac{1}{n h w} \sum_{pixel} \sum_{i = 1}^n \|\mathbf{Q}_i - \mathbf{K}_{f(i)}\|_2^2, 
\end{equation}
where $\sum_{pixel}$ indicates that the term is summed over all the pixel locations of the rendered video. 


\begin{table*}[t]
\centering
\begin{tabular}{l|ccccc|cc}
    & \textit{VLPIPS}$\downarrow$  & \textit{STDerr}$\downarrow$  & \textit{Com.}$\downarrow$  & \textit{Coh.}$\downarrow$  & \textit{LoopQ}$\downarrow$ & \textit{\# Params.}$\downarrow$ & \textit{Render Spd}$\uparrow$ \\ 
\hline
Ours          & \textbf{0.1392} & \textbf{56.02} & \textbf{10.65} & \textbf{9.269} & \textbf{9.263} & 33M-184M & \textbf{140fps} \\
\remark{VBR} & \remark{0.2074} & \remark{82.36} & \remark{12.98} & \remark{11.42} & \remark{11.49} & \remark{300M} & \remark{20fps} \\
loop2D + MTV    & 0.2447 & 118.9 & 11.83 & 9.919 & 9.927 & 33M-184M & \textbf{140fps} \\
loop2D + MPV    & 0.2546 & 117.5 & 11.82 & 9.817 & 9.840 & 2123M & 110fps \\
loop2D + DyNeRF & 0.2282 & 123.7 & 11.93     & 10.23 & 10.27  & \textbf{2M} & 0.1fps  \\
\hline
\end{tabular}
\vspace{-0.2cm}
\caption{Quantitative comparison of reconstruction quality and efficiency. $\downarrow$ ($\uparrow$) indicates lower (higher) is better. Our method produces the best quality and strikes a good balance between the number of parameters and rendering speed. }
\vspace{-0.2cm}
\label{tab:comp}
\end{table*}

\paragraph{Pyramid Training.}
In the implementation, we adopt a pyramid training scheme. In the coarse level, we downsample both the input video and the MTV. The downsampling of the MTV is conducted by downsampling the tiles. We start from the coarsest level with downsample factor $0.24$ and train the MTV representation for 50 epochs. We then upsample each tile by $1.4\times$ and repeat the training step. We show that the pyramid training scheme can improve the generation results.

\section{Experiments}

\begin{figure}
    \centering
    \includegraphics[width=\linewidth,interpolate]{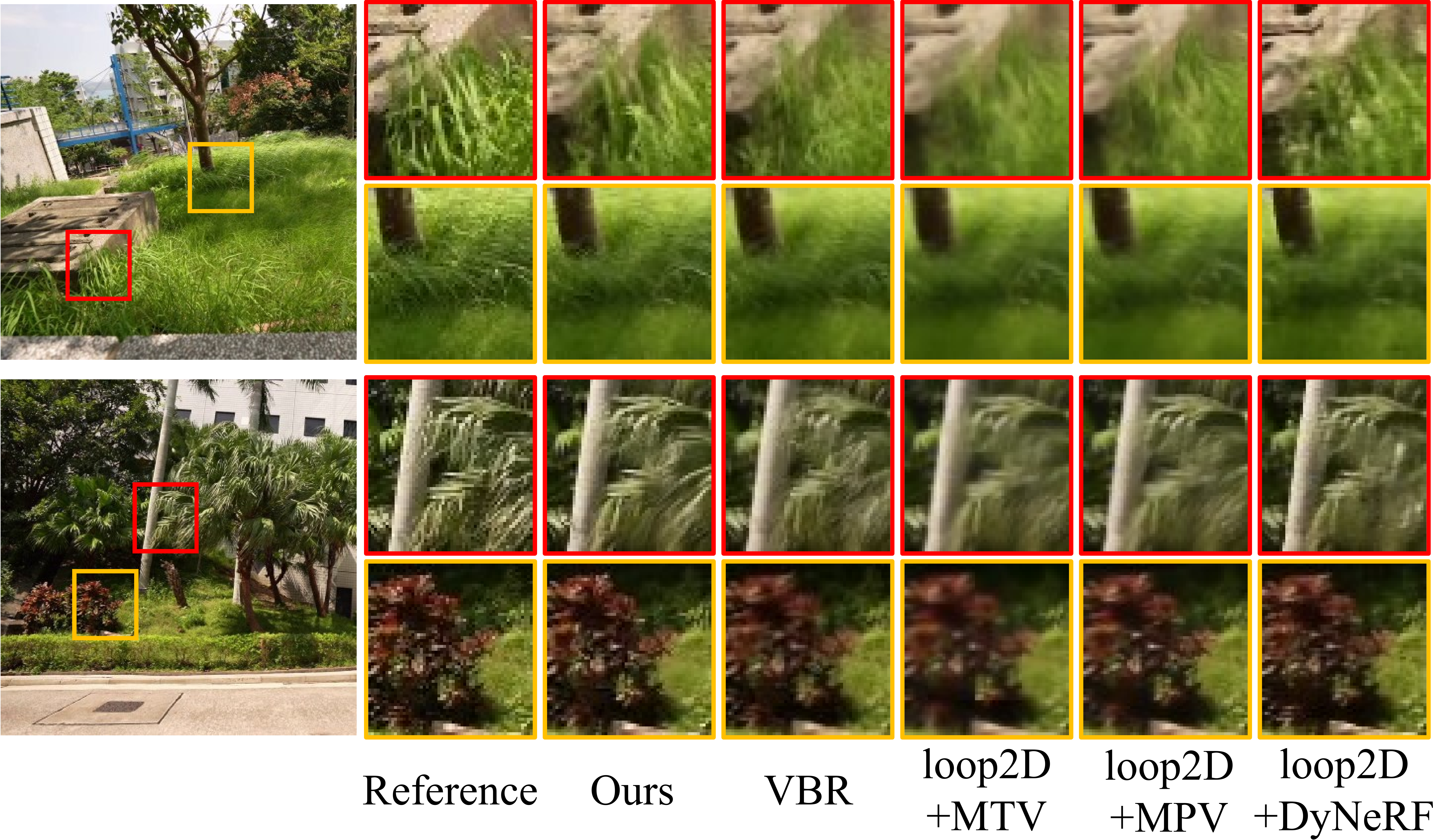}
    \vspace{-0.5cm}
    \caption{Qualitative comparison with other baselines. Our method produces the sharpest results. \lj{the sharpest}}
    \vspace{-0.4cm}
    \label{fig:comp}
\end{figure}

\begin{figure*}
    \centering
    \includegraphics[width=\linewidth]{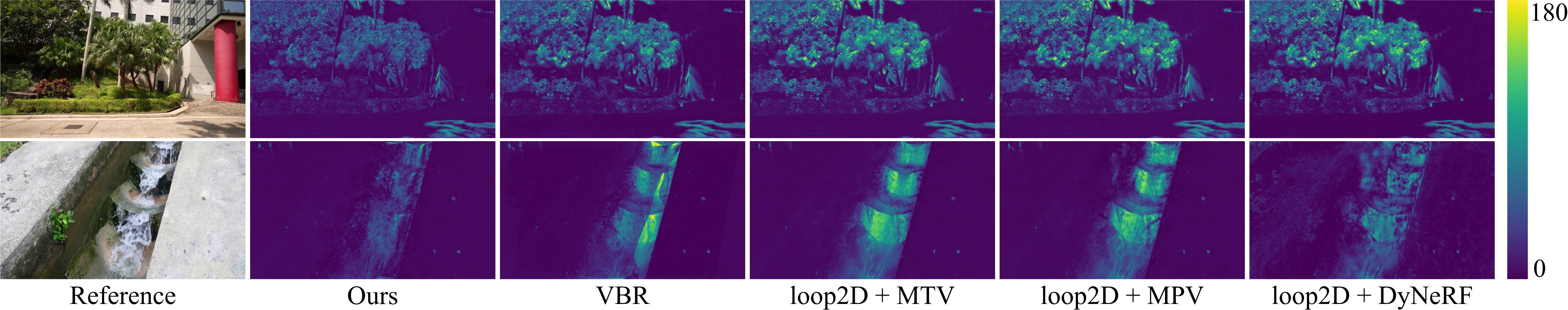}
    \vspace{-0.7cm}
    \caption{We visualize the pixel-wise \textit{STDerr} value for each method. Our method has a lower error, indicating that our approach best retains the dynamism of the scene. We recommend readers watch the supplemental video, where the difference is more noticeable.}
    \vspace{-0.3cm}
    \label{fig:compstd}
\end{figure*}

\subsection{Implementation Details}
We captured 16 scenes for quantitative and qualitative studies. For each scene, we captured 8-10 views in a face-forward manner using a Sony $\alpha 9$ \uppercase\expandafter{\romannumeral2\relax} camera. We captured each view at $25$ fps for 10-20 seconds. We downsample each video to a resolution of $640 \times 360$. Finally, we randomly select one view for evaluation. The others are used for constructing MTVs using the two-stage pipeline. 
In the first stage, we empirically set $\lambda_{tv} = 0.5$ and $\lambda_{spa} = 0.004$. We construct MPI with $D = 32$ layers. 
In the second stage, we let the hyperparameter $\rho = 0$ to guarantee maximum completeness. We extract 3D patches with spatial dimension $11$ and temporal dimension $3$. We construct MTVs with approximately $50$ frames, i.e., 2 seconds. We set the rendering window in each iteration to $h = 180$, $w = 320$ for both stages.

\subsection{Metrics}
For our quantitative study, we synthesize looping videos in test views using the reconstructed 3D video representation and compare the synthetic results with captured target videos. However, we do not have paired ground truth videos since we generate 3D videos with completely asynchronous inputs. Therefore, we adopt several intuitive metrics to evaluate the results in spatial and temporal aspects. 

\noindent\textbf{Spatial Quality.} 
We evaluate the spatial quality of a synthetic frame by computing the \textit{LPIPS} value \cite{misc_lpips} between the synthetic frame with the frame in the target video that is most similar in terms of \textit{LPIPS}. We average the values among all the $50$ synthetic frames, which we denote as \textit{VLPIPS}.

\noindent\textbf{Temporal Quality.}
Given two videos that have similar dynamism, they should have similar color distribution in each pixel location. We measure the temporal quality of the synthetic videos by first computing the standard deviation (STD) of the RGB color at each pixel location of the synthetic video and the target video, resulting in two STD maps of dimension $H \times W \times 3$. We then compute \textit{STDerr} by measuring the MSE between the two maps. 

\noindent\textbf{Spatio-temporal Quality.}
We evaluate the spatio-temporal similarity between the synthetic and target videos following the bidirectional similarity (BDS) \cite{retarget_bds}. We individually report \textit{Completeness} and \textit{Coherence} scores (abbreviated as \textit{Com.} and \textit{Coh.}, respectively) by extracting and finding nearest neighbor 3D patches in two directions. Specifically, for each patch in the target video, we find the closest patches in the synthetic video for \textit{Com.} and vice-versa. We measure the distance of two 3D patches using MSE, and the final scores are the averages of multiple different patch configurations of size and stride. We present the details of the patch configurations in the supplementary material. 

In addition, we use a metric similar to \textit{Coh.} to measure the loop quality (\textit{LoopQ}), which reflects the coherence of the looping video when switching from the last frame back to the first frame. This is achieved by extracting the 3D patches that overlap with the first and last frame, as shown by the blue rectangles in Fig.~\ref{fig:pad}. Other steps remain the same as the \textit{Coh.} score.

\begin{figure*}
    \centering
    \includegraphics[width=\linewidth]{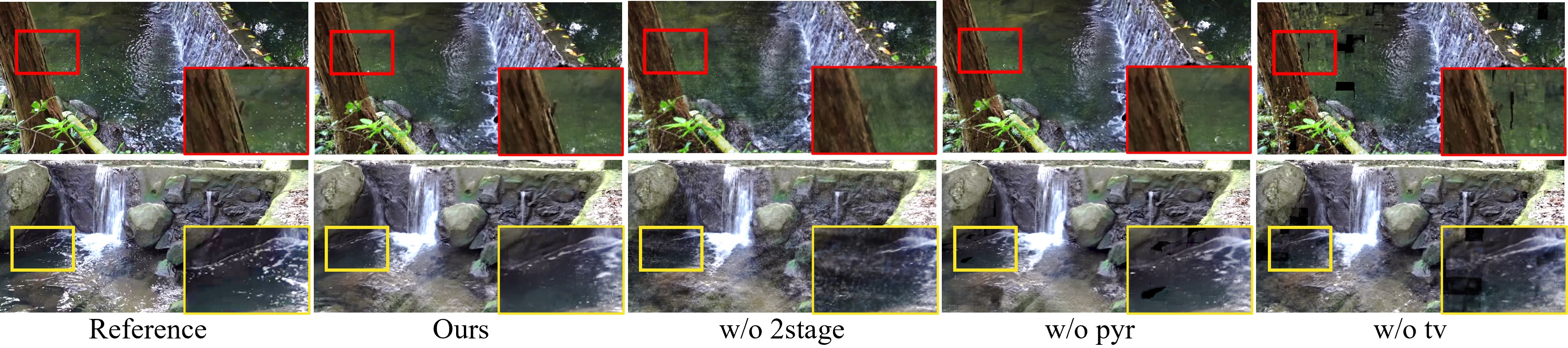}
    \vspace{-0.7cm}
    \caption{Results of our ablations. Our full model produces the fewest artifacts.}
    \vspace{-0.6cm}
    \label{fig:ab4}
\end{figure*}

\subsection{Comparisons}
\ignore{Since there are no prior works that can generate 3D looping videos from asynchronous multi-view videos, we carefully designed several possible baselines to compare with. }
\remark{We first compare with VBR \cite{VBR_loop} by implementing it based on the descriptions in the paper since the code and data are not publicly available. We also compare with straightforward solutions that lift classical 2D looping algorithms to 3D.}
\ignore{We focus on combining the 2D video looping generation method with dynamic novel view synthesis algorithms.} Specifically, we first generate a 2D looping video for each of the input videos using the method of Liao et al. \cite{loop_vid_fast}. And then we construct various scene representations using the 2D looping video and synthesize novel views. We compare with our sparse MTV representation (\textit{loop2D + MTV}), the Multi-plane Video representation (\textit{loop2D + MPV}) and the dynamic NeRF representation \cite{NeRFmvdyn_DynNeRF} (\textit{loop2D + DyNeRF}). 

We compare our method with the \remark{four} \ignore{three} baselines on our captured dataset. We synthesize novel view videos and report \textit{VLPIPS}, \textit{STDerr}, \textit{Com.}, \textit{Coh.} and \textit{LoopQ} metrics in Tab.~\ref{tab:comp}. Our method outperforms other baselines in terms of visual quality, scene dynamism preservation, spatio-temporal consistency, and loop quality. We show the qualitative comparison in Fig.~\ref{fig:comp}. We also visualize the \textit{STDerr} value for each pixel in Fig.~\ref{fig:compstd}, which reflects the difference in dynamism between the synthetic results and the reference. We recommend that readers also see the video results included in the \textit{supplementary material}. 
Note that our method produces the sharpest results, while best retaining the dynamism of the scene. \remark{VBR directly blends inconsistent videos from multiple input views. and the 2D looping baselines fail to consider multi-view information and produce view-inconsistent looping videos. 
} \ignore{The 2D looping baselines does not consider multi-view information and tends to produce view-inconsistent looping videos. 
As a result, the reconstructed scene representations} \remark{As a result, they tend to} blur out spatial and temporal details to compensate for view inconsistencies. 
We observe that \textit{loop2D+DyNeRF} also generates sharper results compared with the other two baselines. This is because DyNeRF conditions on the view direction and tolerates the view inconsistency. However, it performs poorly in maintaining the dynamism of the scene.

\begin{table}[t]
\setlength\tabcolsep{2pt}
\centering
\begin{tabular}{l|ccccc}
\small

& \textit{VLPIPS} $\downarrow$  & \textit{STDerr} $\downarrow$  & \textit{Com.} $\downarrow$  & \textit{Coh.} $\downarrow$  & \textit{LoopQ} $\downarrow$ \\ 
\hline
Ours          & \underline{0.1392} & \underline{56.02} & \textbf{10.65} & \textbf{9.269} & \textbf{9.263} \\
w/o pad       & \textbf{0.1387} & \textbf{55.67} & \underline{10.66} & \underline{9.273} & \underline{9.395} \\
w/o 2stage    & 0.1755 & 67.99 & 11.69 & 9.982 & 10.13 \\
w/o pyr       & 0.1412 & 57.41 & 10.86 & 9.555 & 9.465 \\
w/o tv        & 0.1530 & 56.51 & 11.12 & 9.766 & 9.689 \\
\hline
\end{tabular}
\vspace{-0.3cm}
\caption{Ablations of our method. $\downarrow$ ($\uparrow$) indicates lower (higher) is better. (\textbf{best} in bold, and \underline{second best} underlined)}
\vspace{-0.5cm}
\label{tab:ab}
\end{table}

Additionally, we measure the efficiency of the scene representations using several metrics. We first show the number of parameters (\textit{\# Params.}) of the model to represent a dynamic 3D volume of 50 frames. We evaluate rendering speed (\textit{Render Spd}) at a $360 \times 640$ resolution on a laptop equipped with an RTX 2060 GPU. We present the metrics in Tab.~\ref{tab:comp}. Since the MTV representation varies with different scenes, we report the maximum and minimum values when evaluated in our dataset. \remark{We can see that our method surpasses VBR in \textit{\# Params.} and \textit{Render Spd}.} \ignore{We can see that} Compared with MPV that densely stores the scene parameters in a 4D volume, our sparse MTV representation can reduce the number of parameters by up to 98\%, resulting in a slightly faster rendering speed and much smaller memory and disk usage. On the other hand, despite the surprisingly small number of parameters, the NeRF representation has extremely slow rendering speed. In other words, our MTV representation achieves the best trade-off between the number of parameters and rendering efficiency.


\subsection{Ablation Studies}
\begin{figure}
    \centering
    \includegraphics[width=\linewidth]{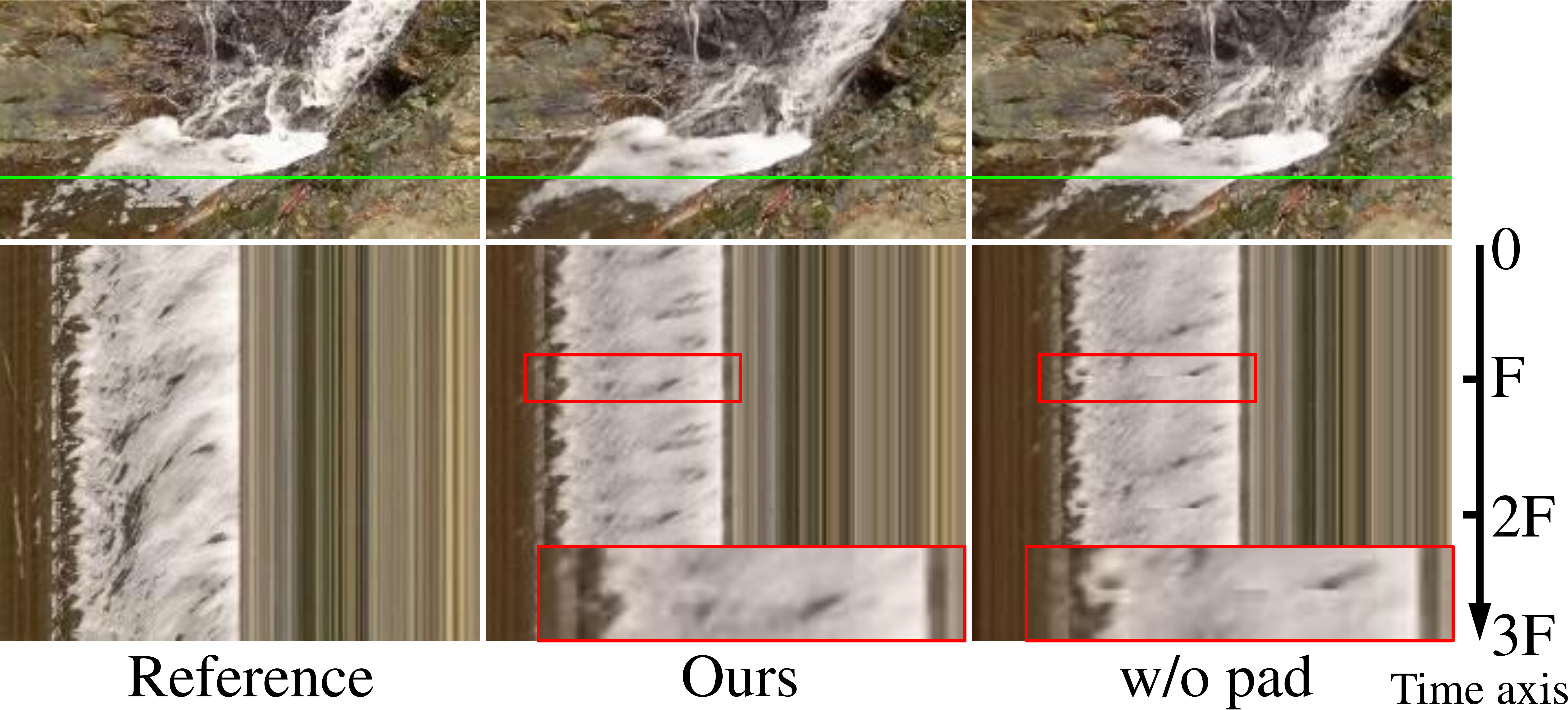}
    \vspace{-0.6cm}
    \caption{Ablations for the padding operation. In the second row, we visualize the temporal coherence by flattening the pixels in the green line along the time axis and repeating $3$ times. Red rectangles highlight the discontinuity produced without the padding operation. We encourage readers to refer to the video results for a clearer demonstration.}
    \label{fig:abpad}
    \vspace{-0.6cm}
\end{figure}

We conducted extensive ablation studies of our method to test the effectiveness of several design decisions in our pipeline by individually removing each component and constructing 3D looping videos from our dataset. We experimented on the following components: the frame padding operation as illustrated in Fig.~\ref{fig:pad} when computing $\mathcal{L}_{loop}$ (w/o pad), the two-stage training pipeline (w/o 2stage), the coarse-to-fine training strategy (w/o pyr), and the TV regularization (w/o tv). The numerical results are shown in Tab.~\ref{tab:ab}, and qualitative results are presented in Fig.~\ref{fig:ab4} and Fig.~\ref{fig:abpad}. We also experimented with different values of $\lambda_{spa}$ and $\rho$ to understand the resulting effect.

\noindent \textbf{Padding Operation.} As shown in Tab.~\ref{tab:ab}, without the padding operation, our method can still produce competitive results in terms of spatial quality and spatio-temporal consistency. It even has better temporal quality. This is because the padding operation adds extra boundary conditions to the optimization, making the optimization more difficult. However, as highlighted in the red rectangles in Fig.~\ref{fig:abpad}, without padding, our method is less prone to generate a properly looping video since it can not guarantee a smooth transition from the last frame to the first frame, leading to a lower loop quality score.

\noindent \textbf{Two-stage Pipeline.} It can be seen from Tab.~\ref{tab:ab} that the two-stage pipeline plays an important role in generating high-quality results. Without the two-stage pipeline, where we directly optimize a dense MPV representation using the looping loss, the MPV easily gets trapped into view-inconsistent results, leading to significant drop in every metric evaluated. 

\noindent \textbf{Coarse-to-fine Training.} Results also show that the coarse-to-fine training scheme produces slightly better spatial and temporal quality than optimizing only on the finest level. This is because the patch-based optimization has a wider perceptual field at the coarse level, leading to a better global solution. Therefore, our full model tends to produce fewer artifacts compared with the \textit{w/o pyr} model. 

\noindent\textbf{TV Regularization.} We find it necessary to apply TV regularization, since the pipeline tends to generate MTVs with holes without this regularization, as shown in Fig.~\ref{fig:ab4}, which greatly affects the visual quality.

\begin{figure}
    \centering
    \includegraphics[width=\linewidth]{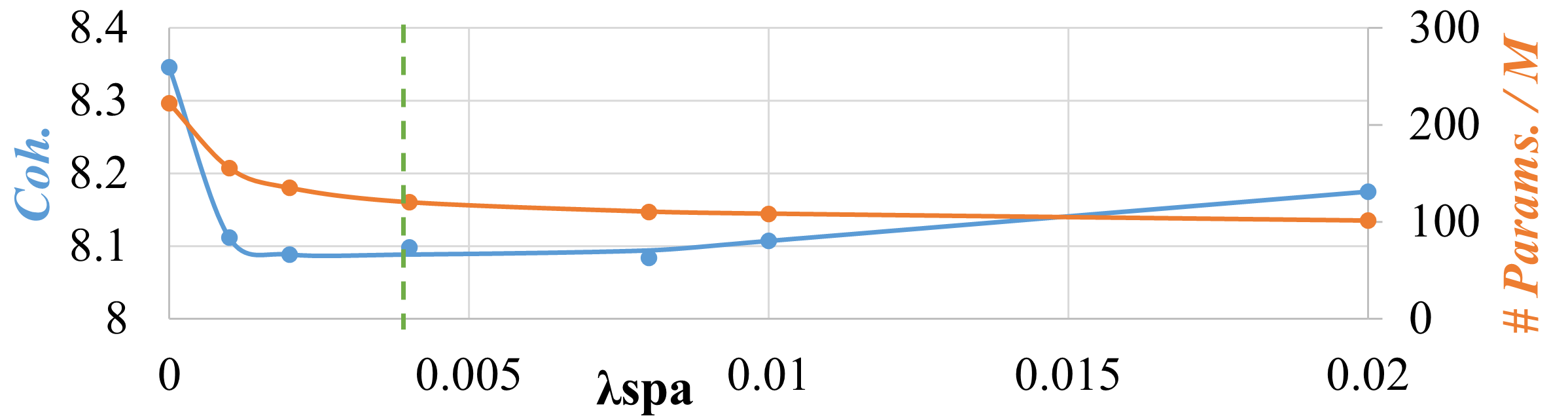}
    \vspace{-0.7cm}
    \caption{The trend of \textit{\color{RoyalBlue} Coh.} score and \textit{\color{Orange} \# Params.} under different $\lambda_{spa}$. The {\color{LimeGreen} green} line is the value we use in all other experiments.}
    \vspace{-0.2cm}
    \label{fig:lambdaspa}
\end{figure}

\noindent\textbf{Weight for $\mathcal{L}_{spa}$.} We experimented on different values of $\lambda_{spa}$ on one scene. We plot the relationship between \textit{Coh.} scores and \textit{\# Params.} with respect to $\lambda_{spa}$. We can see that when $\lambda_{spa} = 0$, the reconstructed MTV is less sparse, 
which degenerates to a dense representation. This makes it harder to optimize and leads to a worse \textit{Coh.} score.
Then \textit{\# Params.} and \textit{Coh.} drop rapidly as $\lambda_{spa}$ grow. However, if $\lambda_{spa}$ is larger than a threshold, \textit{Coh.} increases again, while the improvement on \textit{\# Params.} is less substantial. This is because the excessive sparseness causes the tile-culling process to over-cull necessary tiles, resulting in holes in the rendering results. Therefore, we chose $\lambda_{spa} = 0.004$ (green line in Fig.~\ref{fig:lambdaspa}) in other experiments.

\begin{figure}
    \centering
    \includegraphics[width=\linewidth]{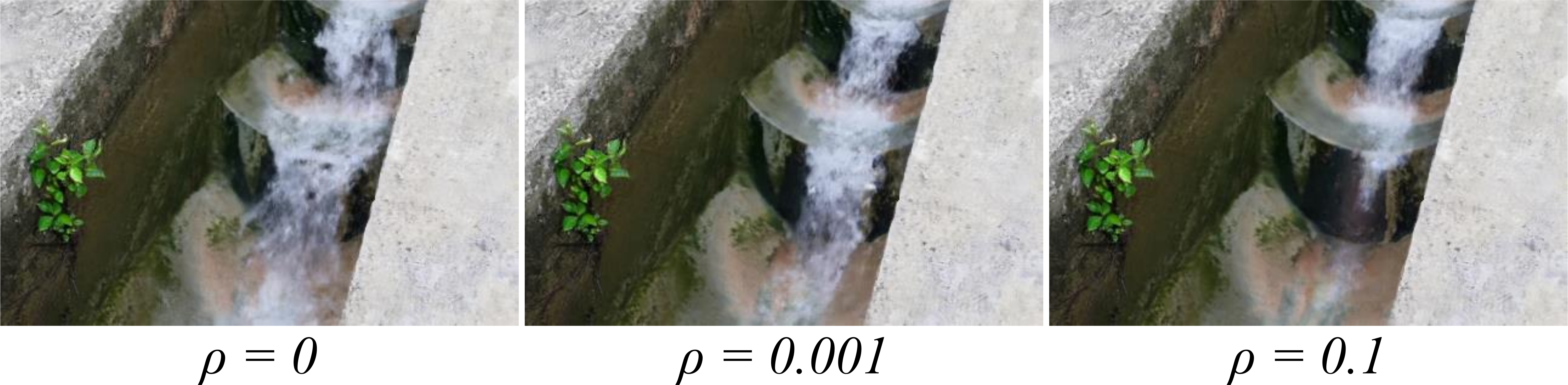}
    \vspace{-0.6cm}
    \caption{Controlling the dynamism by changing $\rho$. }
    \vspace{-0.6cm}
    \label{fig:rou}
\end{figure}

\noindent\textbf{Value of $\rho$.} In the experiments, we use $\rho = 0$ to ensure maximum completeness with respect to the input video. However, we find that by controlling the hyperparameter $\rho$, we could control the degree of dynamism of the reconstructed 3D video. One example is shown in Fig.~\ref{fig:rou}.

\section{Discussion and Conclusion}
\noindent\textbf{Limitations and Future Work.}
Our method comes with some limitations. First, since the MTV representation does not condition on view direction, it fails to model complex view-dependent effects, such as non-planar specular. One possible way to improve the representation is by introducing view-dependency, such as spherical harmonics \cite{sparse_plenoctree} or neural basis function \cite{MPI_Nex}. Another limitation is that we assume the scene to possess a looping pattern, which works best for natural scenes like flowing water and waving trees. However, if the scene is not loopable, our method tends to fail because each view has a completely unique content. This leads to a highly ill-posed problem in constructing a looping video from the asynchronous input videos.

\noindent\textbf{Conclusion.}
In this paper, we propose a practical solution for constructing a 3D looping video representation given completely asynchronous multi-view videos. Experiments verify the effectiveness of our pipeline and demonstrate significant improvement in quality and efficiency over several baselines. We hope that this work will further motivate research into dynamic 3D scene reconstruction.

{\noindent \textbf{Acknowledgements.}} 
The authors from HKUST were partially supported by the Hong Kong Research Grants Council (RGC). The author from CityU was partially supported by an ECS grant from the RGC (Project No. CityU 21209119).

{\small
\bibliographystyle{ieee_fullname}
\bibliography{egbib}
}

\end{document}